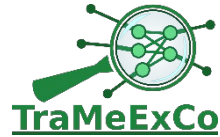

## *Uncovering the Bias in Facial Expressions*

von Jessica Deuschel* [1] [2], Bettina Finzel* [1], Ines Rieger* [1] [2]

[1] Cognitive Systems Group, University of Bamberg, Bamberg
[2] Intelligent Systems Group, Fraunhofer Institute for Integrated Circuits IIS, Erlangen
* All authors have contributed equally to this publication.

## Abstract

Over the past decades the machine and deep learning community has celebrated great achievements in challenging tasks such as image classification. The deep architecture of artificial neural networks together with the plenitude of available data makes it possible to describe highly complex relations. Yet, it is still impossible to fully capture what the deep learning model has learned and to verify that it operates fairly and without creating bias, especially in critical tasks, for instance those arising in the medical field. One example for such a task is the detection of distinct facial expressions, called Action Units, in facial images. Considering this specific task, our research aims to provide transparency regarding bias, specifically in relation to gender and skin color. We train a neural network for Action Unit classification and analyze its performance quantitatively based on its accuracy and qualitatively based on heatmaps. A structured review of our results indicates that we are able to detect bias. Even though we cannot conclude from our results that lower classification performance emerged solely from gender and skin color bias, these biases must be addressed, which is why we end by giving suggestions on how the detected bias can be avoided.

Jessica Deuschel, Bettina Finzel, Ines Rieger

## Introduction

In recent years, Artificial Intelligence (AI) applications have become an increasingly important part of our everyday lives. Netflix recommends relevant films, Amazon offers comparisons to similar products and Siri tells jokes on inquiry. Artificial Intelligence also plays a role in other fields such as security or recruiting procedures. In medicine AI algorithms are also used, for example to detect distinct facial expressions in human faces. These so-called Action Units can help to determine for example whether a patient is in pain. This is especially important when a patient cannot articulate pain by self-assessment.

In order to create such supporting AI solutions, developers usually employ machine learning algorithms which are designed to learn from training data. During the training process, the algorithm finds patterns in the input data and adjusts its operations accordingly and automatically. Some of the best performing machine learning algorithms, such as convolutional neural networks, select automatically which information is most relevant in the input data. In the case of Action Unit detection, the previously mentioned algorithm learns the characteristics of certain expressions in images of human faces and maps them to the corresponding Action Unit, e.g. eye brow lowering. This makes life easier for developers and analysts. However, these algorithms become nontransparent due to the complex decision making processes they rely on: it often remains unclear why they have made a certain decision. In fact, they are black boxes. The consequence of this non-transparency can be incorrectness or unfairness. For example, in their hiring process, Amazon made use of a machine learning algorithm to pre-select applicants for a software developer position, until they realized that the algorithm was biased towards recommending men over women (Dastin, 2018). The algorithm had learned the unequal ratio of both genders in technical jobs from previous years. It had consequently rated the factor 'gender' as highly important for a successful hiring process and effectively discriminated against women as a result.

**Uncovering the Bias in Facial Expressions**

To prevent such errors in the future, it is crucial to verify and ensure fairness of machine learning algorithms. In this contribution we aim to uncover bias resulting from unbalanced training data, taking the example of predicting Action Units. We show that an unbalanced training dataset can lead to a biased model. We attempt to uncover the bias by using interpretability tools, which visually highlight model-relevant parts of the input data and thus increase the transparency of the algorithm's decision. This way, unfairness may be detected and mitigated in future applications.

In the following chapters we want to provide an understanding of both machine and deep learning in general, as well as of the challenging task of Action Unit detection. We put our focus particularly on bias and provide explanations on how it arises and why it is so concerning, especially in machine learning. By manually skewing our data, we show the impact of biased training through the example of Action Unit detection. We also expound the challenges met in detecting and locating the bias. To the best of our knowledge, there is no previous research on uncovering bias for the use case of Action Unit detection.

This contribution is structured as follows: First, a theoretical background is provided explaining machine and deep learning, bias, and Action Units. Next, the data and our experiments are described. In order to test for bias in Action Unit detection, we artificially created skewed, biased datasets that considered only specific subgroups by filtering the data on gender or skin color. For our experiments, we trained several deep learning models on the different dataset splits respectively. The subsequent sectiondescribes our evaluation metricsAfterwards we evaluate the performance of our deep learning models with respect to these metrics and an explainability method. In the final two sections of the contribution we summarize and discuss our results.

Jessica Deuschel, Bettina Finzel, Ines Rieger

## Background

### Machine Learning

For hundreds of years, humans have been fascinated by the idea of creating Artificial Intelligence (AI). The term 'Artificial Intelligence' is not clearly defined and constantly changes along with progress made in the field of computer science. Intelligence in computer science was first defined as the ability to solve rule-based tasks, such as beating a chess champion, and only gradually changed towards more intuitive tasks such as word or image recognition. Such intuitive tasks are especially easy for humans because they can rely on previous experiences. Machine learning, a sub-field of AI, picked up this idea of solving tasks using previous experience. In other words, the machine learning algorithms were designed to 'learn', which here means identifying patterns from 'experience'. Experience is gained from examples, which are provided through training data (Goodfellow et al., 2016).

In the classical sense of machine learning, the developer splits the data into training and test data, extracts features from the training data, and selects a model, for example a linear one. The goal is to find a suitable parameter setting for a model, which minimizes the deviation of the model's predictions from the true values (ground truth), provided by the data. Finding such a parameter setting corresponds to minimizing the error, which is calculated by a given error function, also called loss function.

To give a simple example, consider a regression task of predicting income. The developer can choose useful features from the training data, e.g. the variable 'age' and a plausible model to fit the relation between 'age' and 'income', e.g. a simple linear model, meaning that income increases with age. With these assumptions, the model can learn two parameters: the slope and the start height of this linear relation. 'Learning' in this case means that these parameters are adjusted continuously until a certain loss function - in this case least-squares is common - is minimized. In other words, we find the parameters that lead to the best match between predictions made with our model and the labels of input data. We can also verify the performance of our model by using the test data. In this case, we need to apply the model to the test data and compare the predictions with the ground truth. Assuming that our model performs well, we can process a new data point, e.g. a 42-year-old, and predict the income of this person using our model.

Of course, in a real-world scenario one variable ('age') will not be enough for this task. Other features, for example 'graduation level', 'place of residence' or 'gender' can also play a role in this case. Additionally, most machine learning tasks require not just multiple variables, but also a more complex model. Assuming a linear relationship between the variables and the target ('income') may not apply to children, as they usually have no income. Furthermore, we do not necessarily want to define relevant features beforehand. For example, if our input data consists of images, a manual extraction of relevant features would be challenging. Therefore, the first step, the feature selection, can be included in the modelling process, too. Thus, in the paradigm of our research, the model itself will make a decision about whether the 'color' or 'position' of a pixel (or combination of pixels) in an image is a useful feature for the task at hand by increasing the parameter, also called weight, for this feature. Coming back to our example of income prediction and transferring it to the task of classifying images, we would not need to define relevant features in an image, like information indicating older age. The model itself would extract relevant features automatically, for example wrinkles, in order to predict higher or lower income.

A popular choice for models that do both, feature extraction and prediction for complex tasks, are artificial neural networks. In a neural network both of these steps are completed automatically in the training process. For complex data such as images, we consider deep artificial neural networks, Thus, our research forms part of a sub-discipline of machine learning called deep learning. (Deep) Artificial neural networks received their name from their structure, which resembles the biological brain (Goodfellow et al., 2016). They can process a large amount of data, which is especially useful for images, where the data consists of multiple pixel values. Although (deep) artificial neural networks are structurally complex, their method of operation is in principal the same as that of a simple linear regression. The input data is processed by several mathematical functions that weight the input with adjustable (learnable) parameters. In contrast to linear regression, however, the produced outputs function in turn as inputs for another function, which combines and weights them again. This process is repeated several times. The more functions are chained together (the more so called 'layers' the network has), the deeper it is. If a network has more than 3 layers (input layer + intermediate layers + output layer), it is considered deep. In a nutshell, the input is weighted and recombined in several consecutive operations until we receive a final output representation which can be used to classify the input. For instance, we can use deep artificial neural networks to predict the emotion of a person shown in an image. To train this network, the weights of certain features need to be adjusted just as in other models. In this process, the loss function, which measures the error the model has made, is minimized. As a result the parameters or weights are adjusted such that the model can make a better prediction in the following runs.

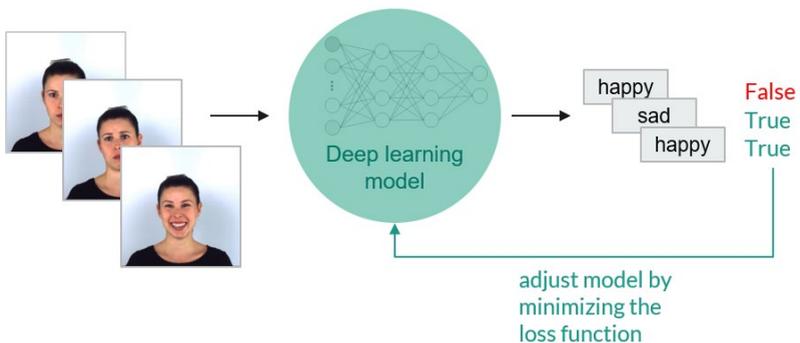

*Figure 1: A schematic overview of the learning process for image classification.*

The described learning process is depicted schematically in figure 1. The input data are images of people showing a particular emotion from which the label or ground truth is extracted. Each image pixel value provides information about the image and is processed by a deep learning model. Based on this input, the model is then given the task to return a class prediction, in this case the emotion depicted in the image. This prediction is either true or false. A loss function calculates the error. In an optimization step it is decided how to change the model weights to arrive at a smaller error. Thus, the model is adjusted by experience, which is called learning. Note that we synonymously use the term 'classifier' for a model performing a classification task. In order to train a deep learning model, a large amount of data is necessary. When we try to teach a model to classify facial expressions, we need several thousand images.

**Bias**

As explained in the previous section, machine learning algorithms heavily rely on data. The data contains all information a model can learn, making its careful selection essential. Skewed data can lead to a biased model, which in our context means that the model may treat certain subgroups, e.g. underrepresented subpopulations, unfairly. Even though the word bias can have several meanings (Dietterich et al., 1995) in our setting it is understood as a skew in the data or model towards certain assumptions which may lead to unfair outcomes. There are several kinds of bias which are not clearly separable. To name a few examples: Mehrabi et al. (2019) distinguishes between omitted variable bias, where critical features that influence the model outcome are missing in the data, and observer bias, which denotes the tendency of humans to see the expected and therefore label data according to their own expectations rather than the relevant feature. Another common bias is the historical bias, a bias reflecting social and historical beliefs in society. An example for the latter is given by Bolukbasi et al. (2016) in their investigation of gender bias in so-called word embeddings. A word embedding – in this case trained on a corpus of 3 million English words from Google News texts – reveals semantic relationships between words in a statement like "London is to England as Berlin to Germany". The same word embedding, however, also created statements such as "man is to doctor as woman to nurse", unveiling outdated concepts in society and gender stereotypes promoted in journalism.

The type of bias we concentrate on is sampling bias (Mehrabi et al., 2019). This kind of bias is introduced by a lopsided data selection that does not represent the population. There are plenty of examples for sampling bias. For instance, the popular dataset ImageNet (Deng et al., 2009) contains only 3% of images from China and India together (Shankar et al., 2017). Shankar et al. evaluate that a model trained on recognizing image content via ImageNet is for example likely to misclassify pictures of Indian bridegrooms while American bridegrooms were mostly classified correctly. Generally speaking, a model trained on a certain population (meaning e.g. a particular geographical location, gender or age group) cannot without issue be deployed on a different population (Kallus et al., 2018). This means that the data used to train models that are to be used for more than one specific population needs to be sufficiently diverse. At this point the question of how to overcome the problem of such sampling bias is posed.

Even in classical machine learning the reduction of bias is a challenge, despite the availability and interpretability of features. Taking a naïve approach – like the complete removal of critical variables such as 'age' or 'gender' – does not prevent unfairness due to alternative correlations (Hardt et al., 2016; Pedreshi et al. 2008). Therefore, input data changes must work in a more sophisticated manner. In an attempt to effect this kind of sophisticated change in input data, Kamiran et al. (2009) reweight the input data using a learned ranking function to lose discriminatory information (e.g. skin color) in the data before training on the modified dataset. Another attempt to overcome bias is by adapting the model. Corbett-Davies et al. (2017) thus investigate algorithmic fairness in machine learning and formulate it as an optimization problem constrained by fairness measures. These fairness measures, however, require an accurate statistical definition. A distinction between these measures can be found in Hardt et al. (2016).

However, for image data and deep learning models, we need different approaches, as image data is too complex and unstructured to apply the kind of fairness measures Corbett-Davis et al. describe. For images, the protected attributes such as 'skin color' are not easily separable from other features because the feature extraction is also not interpretable for humans.

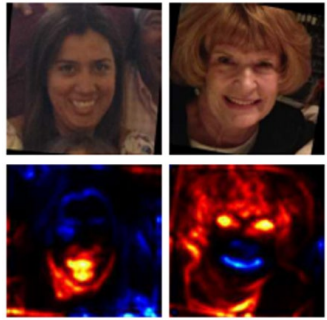

*Figure 2: Layer-wise relevance propagation measuring the relevance of each pixel for age estimation [image from (Lapuschkin et al., 2017)]*

As a first step, bias or unfairness needs to be discovered. Here interpretability and explainability are key. Research in this direction aims to make transparent how and why a machine learning algorithm has come to a certain decision and which features it has considered to be relevant.

As Du et al. (2020) point out, interpretability tools can help to identify bias. The authors describe two ways of identifying bias: from top-down by checking most relevant features that led to a prediction or from bottom-up by slightly changing input features, like changing the skin color of a person, and comparing the change in the prediction. As top-down approach, we can make use of layer-wise relevance propagation (LRP) (Lapuschkin et al., 2017). LRP measures the relevance of each pixel, thus the extent each pixel contributes to the output. The relevance values can be translated into color values which creates a heatmap on the input image showing the most relevant parts in e.g. red and the values contradicting a class in blue. An example from Lapuschkin et al. (2017) is given in figure 2. In this example 'age' is estimated (young vs old). Red pixels contribute to the class 'young', whereas blue pixels speak against it. Here we can identify a bias: The classifier apparently learned that smiling contradicts a person being old. As we find this tool particularly useful, we employ it for our research.

In our research, we mainly focus on identifying bias. However, those interested in related work on mitigating bias may refer to Du et al. (2020) who summarize a few methods on this matter and Wang et al. (2020) who benchmark bias mitigation techniques. For a practical example of bias identification and mitigation, see for example the previously mentioned word embeddings in Bolukbasi et al. (2016). For a comprehensive overview of methods, IBM Research developed the AI Fairness 360 Open Source Toolkit[1], a set of fairness metrics for datasets and models, explanations for these metrics, and algorithms to mitigate bias in datasets and models (Bellamy et al., 2018).

Other research has put the focus on bias and Action Units, the use case this paper focuses on. In Kilbride et al. (1983) the authors conducted a survey on ethnic bias in emotion recognition where Action Units served as ground truth label. In this survey, human bias, as opposed to machine learning algorithm, was the topic of investigation. More directly related to our research field is Xu et al. (2020). In their research, the authors investigate bias in facial expression recognition, albeit without the consideration of Action Units. Also loosely related is the work of Serna et al. (2020) who examine and mitigate bias in face recognition. For the

---

[1] The IBM AI Fairness 360 Open Source Toolkit and more information on its uses can be found here: https://aif360.mybluemix.net/.

**Uncovering the Bias in Facial Expressions**

research presented here, we build on this previous research to uncover bias in classification of facial expressions based on Action Units.

## Action Units

Action Units are individual facial expressions, which are distinguished and defined in the manual of the Facial Action Coding System (Ekman et al., 1987). The manual describes the occurrence and possible combinations of different Action Units.

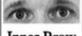

*Figure 3: A subset of Action Units from the Facial Action Coding System (Yin et al., 2017).*

Figure 3 lists a subset of existing Action Units. Action Units can occur in the lower parts of the face or the upper face. For example Action Unit 1, which is the inner brow raiser, belongs to the Action Units in the upper face. Action Unit 10, which corresponds to raising the upper lip, occurs in the lower face.

Jessica Deuschel, Bettina Finzel, Ines Rieger

Interpreting facial expressions correctly is especially relevant in the clinical context. Combinations of Action Units can be indicators for emotions as well as pain and may help in the assessment of the state of a patient who cannot communicate the intensity of pain he or she feels. Therefore, reading emotions and pain in the patient's face, which is to say predicting and interpreting Action Units correctly, can be helpful to decide about appropriate treatment. However, systems have to be trained carefully, since an unwanted sampling bias can be easily induced, producing severe consequences for individual patients or even groups of patients.

Action Units can be expressed very differently, as their intensity and frequency vary. The appearance of Action Units can be influenced by age (for example through more wrinkled faces) and health (presence of scars etc.). In the following, we explore sampling bias for gender and skin color and examine these two research questions:

1. Will a bias resulting from skewed data manifest in the model's predictive performance?
2. Will this bias be observable in visual explanations?

The hypothesis underlying our research questions is that the neural network will analyze correct regions for male participants and irrelevant or even wrong regions for female participants, when its predictive performance is worse for female compared to male participants, and vice versa. The same is assumed for participants with light skin versus participants with dark skin. In the following section, we describe the experiments conducted in order to evaluate our research questions.

## Experiments

We conducted experiments for evaluating bias by training on a skewed dataset distribution with imbalanced gender and skin color representation.

**Uncovering the Bias in Facial Expressions**

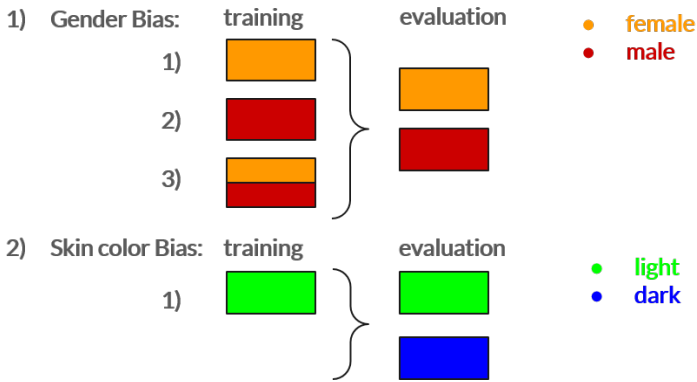

Figure 4: Our conducted experiments with skewed datasets.

We split our first experiment for evaluating gender bias in three parts: We trained a model on 1) only female subjects, 2) only male subjects and 3) on male and female subjects. The number of subjects (37) for each of these training sets is the same for all experiments, thus making the number of images and annotated Action Units comparable. For experiment 3) with mixed genders, the subjects are a random subset of the female subjects of experiment 1) and the male subjects of experiment 2). Each of these gender experiments are evaluated on the same testing set of 10 subjects of each gender. The number of training and testing subjects is limited by the male subjects, where we have 47 subjects in total.

In our second experiment, we train a model on a dataset with imbalanced skin color distribution. Since we have very few data on dark skin color, merely 19 subjects, we could only make one experiment, in which we train a model on people with relatively light skin and evaluate on a testing set of people with relatively dark skin. We thus trained a model on 104 'light-skinned' subjects and evaluated this model on 19 'dark-skinned' subjects. In our experiment we looked at gender and skin color from the perspective of machine learning, and thus as a collection of facial characteristics that possibly, but not necessarily, correlate with the racial identities and social positioning that are commonly associated with them.

## Data

We used the recent Actor Study (Seuss et al., 2019) dataset and the widely used benchmark dataset CK+ (Lucey et al., 2010) for training and evaluation. The Actor Study (Seuss et al., 2019) contains sequences of 21 actors, each of them filmed for about 68 minutes from different views. This results in 4,337 frames per subject on average. The recording took place in a lab setting, so there is no change in background and lighting. The actors were asked to display specific Action Units and to react to scenarios and enactments. Experts annotated each frame of the sequences with a set of Action Units. For our experiments, we used the center view of the low speed camera. This dataset was recently published and will be made publicly available for commercial research.

The CK+ (Lucey et al., 2010) dataset contains 593 videos of 123 subjects in a lab setting. There are 87 frames on average available for each subject. Each sequence shows a facial expression - one of the basic emotions - from neutral to strong. All expressions are acted out. Experts then annotated a set of Action Units for the whole sequence, and not on a video frame level. For this, they only looked at the last frames, where the expression is the strongest. This dataset is for non-commercial research purposes only.

## Data Pre-Processing

The occurrence of Action Units is naturally imbalanced in datasets, since Action Units do not occur with the same frequency for different facial expressions. The more input data provided, the better deep learning models perform. For training we therefore selected only Action Units that have at least 8,000 occurrences (see figure 5). Occurrence here means the display of the Action Unit on a face.

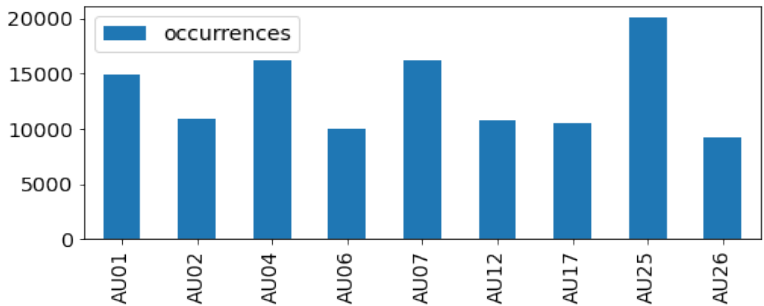

*Figure 5: The number of occurrences per Action Unit of the whole training and evaluation dataset.*

Furthermore, neural networks perform better when the classes are balanced. There are techniques to balance multi-label multi-class datasets (Rieger et al., 2020), but in order to not skew the amount of images in an uncontrolled manner per gender or skin color, we do not balance our data. We trained our model on color images, as these provide additional information. For each experiment, we split the training data in a training and validation set and evaluated the trained model on the prepared testing dataset.

**Training**

As proposed in the paper of Pahl and Rieger (2020), we use a neural network with a VGG16 architecture (Simonyan et al., 2014) that is pre-trained with the ImageNet dataset (Deng et al., 2009) (see figure 6). The ImageNet dataset used for pre-training has 1,000 classes with various objects. Therefore, these pre-trained models have already some idea of how to 'read' images and can be adapted to new domains by fine-tuning them. We fine-tune this pre-trained model in our experiments on our Action Unit training dataset.

Jessica Deuschel, Bettina Finzel, Ines Rieger

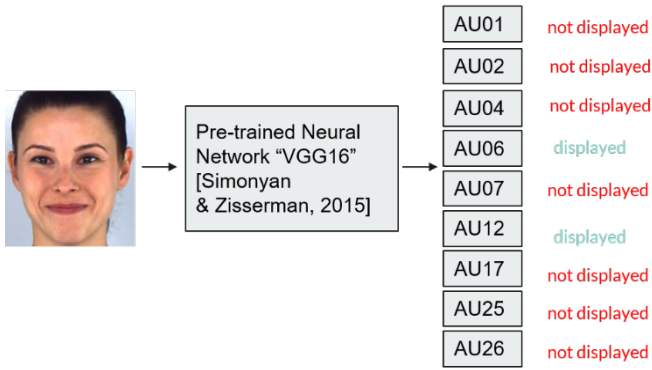

*Figure 6: Our training pipeline, where we adapt the pre-trained neural network to our Action Unit domain.*

In contrast to Pahl and Rieger (2020), we fine-tuned only the last layers - those that are fully connected. For further information on hyperparameter values, pre-processing, the loss function, and architectural details, please refer to Pahl and Rieger (2020).

**Performance Measurement**

Next to a quality evaluation using heatmaps, as explained previously, predictive performance was examined. One measure that can be applied, is accuracy. This measures how many instances from a dataset have been classified correctly. Accuracy ranges from 0 to 100%. However, it may not tell us whether our model is a sophisticated predictor. Let us take for instance an imbalanced dataset with seven instances belonging to a class (e.g. Action Unit 1) and three not belonging to that class (e.g. Action Unit 2). A model which simply classifies all instances as positive (i.e. Action Unit 1), without taking into account any characteristics of the data and thus making a random decision for one label, achieves an accuracy of 70% on this dataset (see figure 7). Here, the model did not learn anything, since its choice of labels was random. We could do the same for the negative class.

**Uncovering the Bias in Facial Expressions**

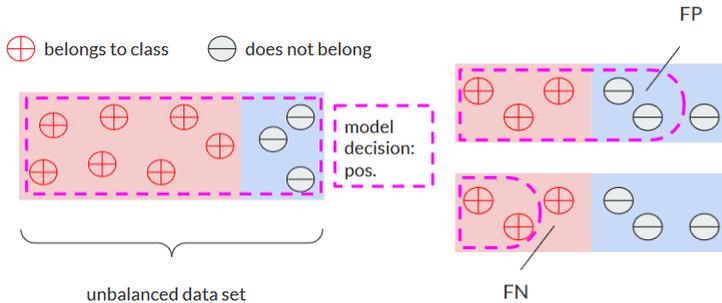

Figure 7: *The relationship between classifier accuracy and imbalanced class labels on the left and illustrates false positives (FP) and false negatives (FN) on the right.*

In the first case, we have false positives (meaning negative examples that were mistakenly classified as positive). In the second case, we have false negatives (meaning positive examples that the classifier missed). False positives (FP) and false negatives (FN) are illustrated in figure 7. A better measure in comparison to accuracy is the F1 score below. It takes precision and recall into account (see figure 8). Precision measures how well false positives are avoided by the algorithm. The recall measures how well false negatives are avoided.

$$F_1 = 2 \cdot \frac{\text{precision} \cdot \text{recall}}{\text{precision} + \text{recall}}.$$

Figure 8: *The formula for computing the F1 score*

In the following section, we present our results, evaluating our trained model qualitatively by means of heatmaps and quantitatively on predictive performance.

**Results**

Table 1 shows our first experiments, where we evaluate gender bias. As we recall from figure 4, we trained on 1) only female subjects, 2) only male subjects, and 3) subjects of both genders, and we evaluated on female and male subjects. A weighted F1 macro score takes the number of samples per class into account, so the F1 score per class is weighted on the number of samples for this class.

| Trained on | Tested on female | | Tested on male | |
|---|---|---|---|---|
| | F1 macro | weighted F1 macro | F1 macro | weighted F1 macro |
| female | 0.55 | 0.55 | **0.56** | **0.58** |
| male | **0.44** | **0.45** | 0.55 | 0.59 |
| mixed | 0.54 | 0.55 | 0.60 | 0.63 |

*Table 1: The results of the experiments in which a potential gender bias is evaluated.*

We can see from the results that a gender bias is observable, as the results on the testing set change depending on which gender we train the model on. When training on male subjects, the results meet our expectation, as the F1 macro is lower when the trained model is evaluated on female subjects (0.44) than when evaluated on male subjects (0.55). Our expectation is also met when evaluating the model trained on subjects of both genders: The F1 macro for female testing subjects (0.54) is as high as when evaluating the model trained on female subjects (0.55). Furthermore, the F1 macro on the male testing subjects (0.60) is almost as high as when evaluating the model trained on male subjects (0.55).

Our expectations are not met, however, when the model is trained on female subjects only, which shows in fact that the F1 macro scores are almost equal when the model is evaluated on the male testing subjects (0.56) and on the female testing subjects (0.55). We can put these results in perspective when comparing the F1 macro on the female testing subjects (0.55) from this experiment with the F1 macro on the female testing subjects from the experiment, where the model is trained on male subjects (0.44): The result is higher in the former experiment as expected.

**Uncovering the Bias in Facial Expressions**

But when we look at the male testing subjects and compare the experiment in which the model was trained on female subjects (0.56) with the experiment in which the model was trained on male subjects (0.55), the results are almost the same. We must therefore assume that there are additional dependent variables in the dataset that influence the results.

Overall, we see for the gender experiment that the results for a successful recognition of the Action Units on the female testing subjects are always lower than on the male testing subjects. We hypothesize that there might be more diverse features in the female dataset than in the male dataset, such as more variance in appearance. Testing for this hypothesis is not trivial, however, as one must first define the scope of features that are taken to characterize men and women respectively, e.g. hairstyle. Furthermore, if these features are not labeled in the dataset, a model has to be trained to recognize them. We leave this to future work.

Figure 9 shows sample heatmaps computed of female subjects that were created based on the LRP method. The heatmaps indicate how a model trained on male subjects 'looks' at these images of female subjects, when predicting Action Unit 1, the inner brow raiser:

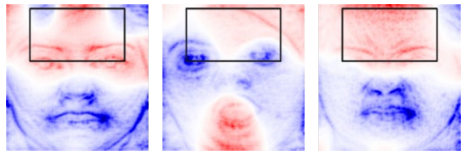

*Figure 9: How a model trained with male subjects 'sees' images of female subjects when predicting the Action Unit 1 – the inner brow raiser.*

In all three images, Action Unit 1 is present in the face and the model predicted it correctly as present. The approximate position of the inner brow raiser is marked with a black rectangle. In the following, we describe the heatmaps and propose explanations for the results:

- In the first image (left) the person has a hairstyle with bangs that cover part of the forehead. These areas on the forehead are blue in the heatmap, which means that the trained model sees the strands of hair as opposing features that hinder the classification of Action Unit 1, the inner brow raiser. By contrast, the model sees the red area on the forehead as important for this Action Unit class. We can thus assume that the model is confused by the hairstyle.
- In the second image (middle), the model does not only look closely at the forehead, which is the expected area of the relevant Action Unit class, but also at the hair and chin/mouth area. We can thus again assume that the model learned wrong statistical correlations: a correlation between the chin/mouth area and the forehead for recognizing Action Unit 1. This can also occur due to too little training data.
- The third image (right) shows a prediction we would expect: the forehead is important for this class; everything else is blue.

Table 2 shows our second experiment, where we explore a possible skin color bias. As expected, the F1 macro on the testing subjects with light skin color is higher (0.63) than the result on the testing subjects with dark skin color (0.55), when evaluated with a model trained on light skin color. Therefore, the skewed skin color training dataset here results in a biased and unfair model.

| Trained on light skin color | Tested on light skin color | | Tested on dark skin color | |
|---|---|---|---|---|
| | F1 macro | weighted F1 macro | F1 macro | weighted F1 macro |
| | **0.63** | **0.69** | **0.55** | **0.62** |

*Table 2: The results of the experiments, in which a potential skin color bias is evaluated.*

**Uncovering the Bias in Facial Expressions**

Figure 10 shows three sample heatmaps computed on our testing subjects with darker skin:

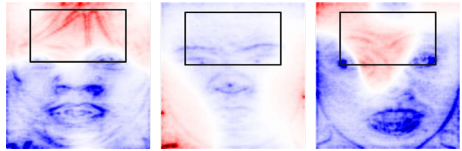

*Figure 10: How a model trained with light-skinned subjects 'sees' images of dark-skinned subjects, when predicting the Action Unit 1 - the inner brow raiser.*

The heatmaps show how a model trained with light-skinned faces 'sees' dark-skinned faces, when predicting the Action Unit 1, the inner brow raiser. In the following, we describe the heatmaps and propose explanations for the results:

- We can see again in the first heatmap (left) how a certain hairstyle affects or confuses the trained model. The model predicted that the Action Unit 1 would be present in this image, although it is not. Probably there were not enough training images featuring this particular hairstyle in order to avoid the bias.
- For the second image (middle), the trained model predicts correctly that there is no Action Unit 1 present in this image. However, we can see that the model looks at the background to support its prediction, which can indicate a background bias, probably induced by the monotone lab background.
- For the third image (right) the model correctly predicts the presence of Action Unit 1. The heatmap highlights the forehead area and especially the wrinkles, as we would expect. The rest of the face is considered by the trained model as opposing to the class, which is correct.

Jessica Deuschel, Bettina Finzel, Ines Rieger

## Discussion

In this paper, we raised and experimentally examined two research questions. We focused on answering whether introducing an imbalance in gender and skin color into the training dataset would result in a bias. We assumed that the bias would manifest in the model's predictive performance, since non-representative sampling corrupts performance. We further hypothesized that a bias would be observable in visual explanations, presented as heatmaps.

Our quantitative results show that the predictive performance on female subjects was indeed reduced when the model had been trained only with male subjects, in comparison to training with a mixed dataset or only on female subjects. We can furthermore observe that the predictive performance of a model trained only on female subjects has a better prediction outcome on male subjects, compared to the all-male and mixed dataset case. For skin color, the model's performance decreases on test subjects with darker skin compared to subjects with lighter skin. We can thus conclude that a bias arose from skewing the data with respect to gender and skin color. Otherwise, the predictive performance would have stayed approximately the same among the different settings.

With respect to our qualitative analysis on the appearance of visual explanations, we assumed that the neural network would consider Action Unit-relevant image regions for male participants and irrelevant regions for female participants, when its predictive performance is worse on the test data for female subjects compared to male subjects. The same outcome was assumed for the experiment focusing on the participants' skin color. Although the model seems to concentrate on specific areas in some cases (such as the forehead when the prediction was correct or the background of the image when the prediction was wrong, as in the example of the inner brow raiser), our main observation is that the heatmaps produced are hardly homogenous.

We visually observed the potentially biased heatmaps. We found that regions irrelevant to Action Units have been highlighted, but we cannot

Uncovering the Bias in Facial Expressions

conclude in general that these are features specific of participants of a certain gender or with a certain skin color.

At this stage, our second research question cannot be answered in a satisfactory manner, since we have not yet performed an extensive quantification of the appearance of the heatmaps. This must remain the object of future work. In order to sufficiently evaluate the heatmaps, a metric that calculates the aggregation of positive relevance in Action Unit-specific facial regions needs to be applied. Additional annotations of these regions would have been required, but they were not available at the time of this work.

With our experiments, we aimed to show that a sampling bias can reduce the performance of a model. Our results indicate that there is an influence, which is why mitigating bias in datasets is an important step towards more objectivity in machine learning algorithms. In order to reduce bias, using more diverse datasets can improve the model's ability to generalize. Nevertheless, we do not think that it would be reasonable to try to eliminate any kind of bias completely. Bias is at the core of learning as it allows for abstraction from individual examples. There is no absolute knowledge in the world, so that every abstraction, being a reduction of information, can be an undesired or desired bias. Eliminating the bias completely would remove the model's ability to perform an abstraction from examples. The model would then simply have to memorize each example in the given representation. Therefore, removing all bias is not desirable. Eliminating specific discriminatory bias, however, improves generalizability.

We further want to point out that developers and deployers of machine learning algorithms and models respectively can end up conserving or even promoting stereotypes if they take the input data for training from a biased source. Such a bias (which may come from societal biases at large) would be inherent to the learned model and thus base its decisions on prejudices and assumptions similar to those present in the human data collectors. It is therefore desirable to take data from different or diverse sources. This includes involving multiple and diverse human experts in the process of instance labelling for the training data in an effort to integrate various viewpoints.

With respect to the step of decision making, we consider increasing transparency of machine learning and finding measures that ensure transparency, objectivity, and fairness to be crucial for the evaluation of a model's performance. A central question is therefore, whether the model's output should help to achieve the best possible results – here one that is as fair as possible – or whether it is enough if a model is as good as the world that has produced it. Nevertheless, without transparency, objectivity and fairness measures, bias evaluation and decision making would be limited. In our future research we therefore aim to proceed in this direction.

## Conclusion & Future Work

In this paper, we show by means of Action Unit detection that an induced sample bias in the training data can lead to performance reduction on the underrepresented population. We considered gender and skin color bias that could lead to discrimination if a model is not representative enough. We built several models, each trained on the same number of subjects, but one specific subgroup respectively: only male, only female, both genders, only participants with light skin color and participants with all skin colors. Our gender analysis reveals that the performance on female subjects drops considerably when training on exclusively male subjects in comparison to training on both genders. We detect the same results for the skin color analysis: The score for dark skin improves when considering subjects with different skin colors in the training data as opposed to only subjects with light skin. Surprisingly however, inverting the experiment, thus training on only female subjects, does not result in a performance reduction for the male subgroup. We suspect, therefore, that another kind of bias is present in the data. We speculate that creating subsets has reduced diversity in the data, yielding inflexible models that do not know enough variation. This could be avoided using a larger, more diverse and balanced dataset. For skin color, this reverse analysis was not possible due to the limited amount of data for subjects with darker skin color.

In addition to the performance comparison, we investigated bias applying the explainability method LRP. We discovered that for the respective underrepresented group the models sometimes learned to consider spurious correlations, like when they looked at parts of the background in addition to the face, and also became confused by unexpected attributes, such as different hairstyles. This also suggests the importance of augmenting diversity in the dataset. However, evaluating these heatmaps is dependent on personal interpretations. In order to evaluate the quality of the produced explanatory heatmaps objectively, a domain-specific measure could be applied, as presented in Rieger et al. (2020). For the verification of the model's performance, bounding boxes are defined for Action Unit relevant regions and the aggregated positive relevance inside and outside of the bounding boxes computed. Overall, with this research, we hope to contribute to a heightened state of awareness and comprehension for bias in deep learning and contribute particularly to research on Action Unit detection by presenting ways to uncover bias with respect to this task.

## Acknowledgments

We would like to thank the women's representatives of the University of Bamberg for the competence- and network-oriented support provided during the colloquium FORSCHEnde FRAUEN. The experiments and results presented here were partly generated within the research project Transparent Medical Expert Companion and financially supported by the BMBF (FKZ 01IS18056 A/B, TraMeExCo). This work was further supported by the Bavarian Ministry of Economic Affairs, Regional Development and Energy through the Center for Analytics Data

**Jessica Deuschel, Bettina Finzel, Ines Rieger**

Applications (ADA-Center) within the framework of "BAYERN DIGITAL II" (20-3410-2-9-8). Furthermore, we would like to thank Dominik Seuss for his support, Jaspar Pahl for his constructive feedback as well as his help in the preparation of the experiments and Sebastian Lapuschkin for the image material provided to showcase age bias. We also thank Sarah Stenz for proofreading our contribution.